\documentclass[letterpaper]{article} 
\usepackage{aaai25}  
\usepackage{times}  
\usepackage{amssymb}
\usepackage{pifont}
\usepackage{courier}  
\usepackage{booktabs}
\usepackage{times}  
\usepackage{helvet}  
\usepackage{amssymb}
\usepackage{amsmath}
\usepackage[capitalise]{cleveref}
\usepackage{float} 
\usepackage{graphicx}
\usepackage{courier}  
\usepackage[hyphens]{url}  
\usepackage{natbib}  
\usepackage{caption} 
\usepackage{algorithm}
\usepackage{algorithmic}

\usepackage{newfloat}
\usepackage{listings}
\DeclareCaptionStyle{ruled}{labelfont=normalfont,labelsep=colon,strut=off} 
\floatstyle{ruled}
\newfloat{listing}{tb}{lst}{}
\floatname{listing}{Listing}
\title{FBRT-YOLO: Faster and Better for Real-Time Aerial Image Detection}

\author{
    Yao Xiao, Tingfa Xu\thanks{Corresponding authors.}, Yu Xin, 
    Jianan Li$^\ast$} 
\affiliations{
    Beijing Institute of Technology\\

    \{3220220586, ciom\_xtf1,3220220587,lijianan\}@bit.edu.cn\\

}

\usepackage{bibentry}

\begin{document}

\maketitle

\begin{abstract}
Embedded flight devices with visual capabilities have become essential for a wide range of applications. 
In aerial image detection, while many existing methods have partially addressed the issue of small target detection, challenges remain in optimizing small target detection and balancing detection accuracy with efficiency.
These issues are key obstacles to the advancement of real-time aerial image detection.
In this paper, we propose a new family of real-time detectors for aerial image detection, named FBRT-YOLO, to address the imbalance between detection accuracy and efficiency. Our method comprises two lightweight modules: Feature Complementary Mapping Module (FCM) and Multi-Kernel Perception Unit (MKP), designed to enhance object perception for small targets in aerial images.
FCM focuses on alleviating the problem of information imbalance caused by the loss of small target information in deep networks. It aims to integrate spatial positional information of targets more deeply into the network, better aligning with semantic information in the deeper layers to improve the localization of small targets.
We introduce MKP, which leverages convolutions with kernels of different sizes to enhance the relationships between targets of various scales and improve the perception of targets at different scales.
Extensive experimental results on three major aerial image datasets, including Visdrone, UAVDT, and AI-TOD, demonstrate that FBRT-YOLO outperforms various real-time detectors in terms of performance and speed. Code is will be avaliable at https://github.com/galaxy-oss/FCM.
\end{abstract}

\section{Introduction}

Recent advancements in deep neural networks have significantly improved object detection in low-resolution natural images~\shortcite{wang2022anchor,wang2023yolov7}. However, these methods struggle with efficiency and accuracy on high-resolution aerial images, especially in resource-constrained flight equipment. Key challenges include: i) detecting objects that are small or obscured by backgrounds in aerial images, and ii) balancing accuracy with real-time detection requirements on devices with limited computational resources.

\begin{figure}[t] 
    \centering
    \includegraphics[width=1\linewidth]{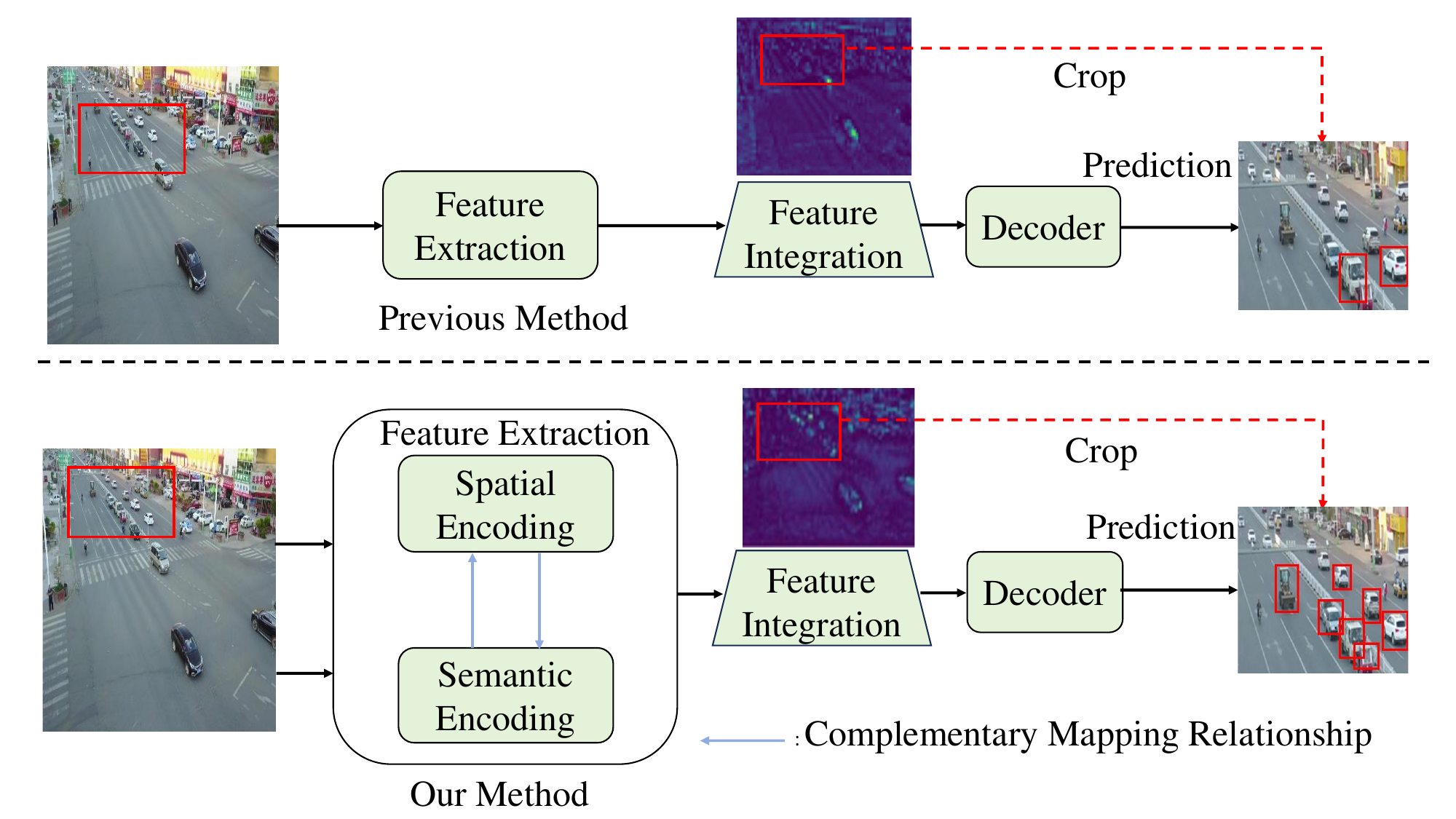}
    \caption{The previous method overlooked the embedding of spatial information in deeper layers of the backbone network during feature extraction, leading to spatial semantic inconsistencies. Our method aims to transfer shallow spatial location information into deeper layers of the network during the feature extraction process, thereby enhancing the expression of semantic information.}
    \label{1}
\end{figure}

To improve small object detection, increasing image resolution~\shortcite{jianan1,wang2020deep} is common but adds computational burden, hampering real-time performance. A key challenge is the mismatch between low-resolution semantic information from deep networks and high-resolution spatial information from shallow networks.
Feature pyramids~\shortcite{FPN} address this by integrating deep and shallow features, enhancing small object localization and multi-scale feature expression while improving computational efficiency. 
However, as shown in \cref{1}(a), backbone networks still struggle with integrating and preserving shallow information, leading to feature mismatch issues.

To address the challenges associated with object detection in aerial images, we aim to achieve a more effective network design that meets the requirements for both accuracy and efficiency in real-time aerial image analysis. In this paper, we propose a novel network that includes two lightweight modules: the Feature Complementary Mapping Module (FCM) and the Multi-Kernel Perception Unit (MKP).
Firstly, to alleviate information imbalance within the backbone network and promote better integration of semantic and spatial location information, we introduce the Feature Complementary Mapping Module (FCM). FCM implicitly encodes the target's spatial location information into high-dimensional vectors, guiding the complementary learning of spatial and channel information across different stages of the backbone network. This facilitates the fusion of shallow spatial location information with deep semantic information, enhancing the consistency of spatial and semantic representations. This approach helps transfer shallow spatial location information to deeper layers of the network, improving feature alignment and enhancing the localization of small objects, as shown in \cref{1}(b).
\begin{figure}[t] 
\centering
\includegraphics[width=0.97\linewidth]{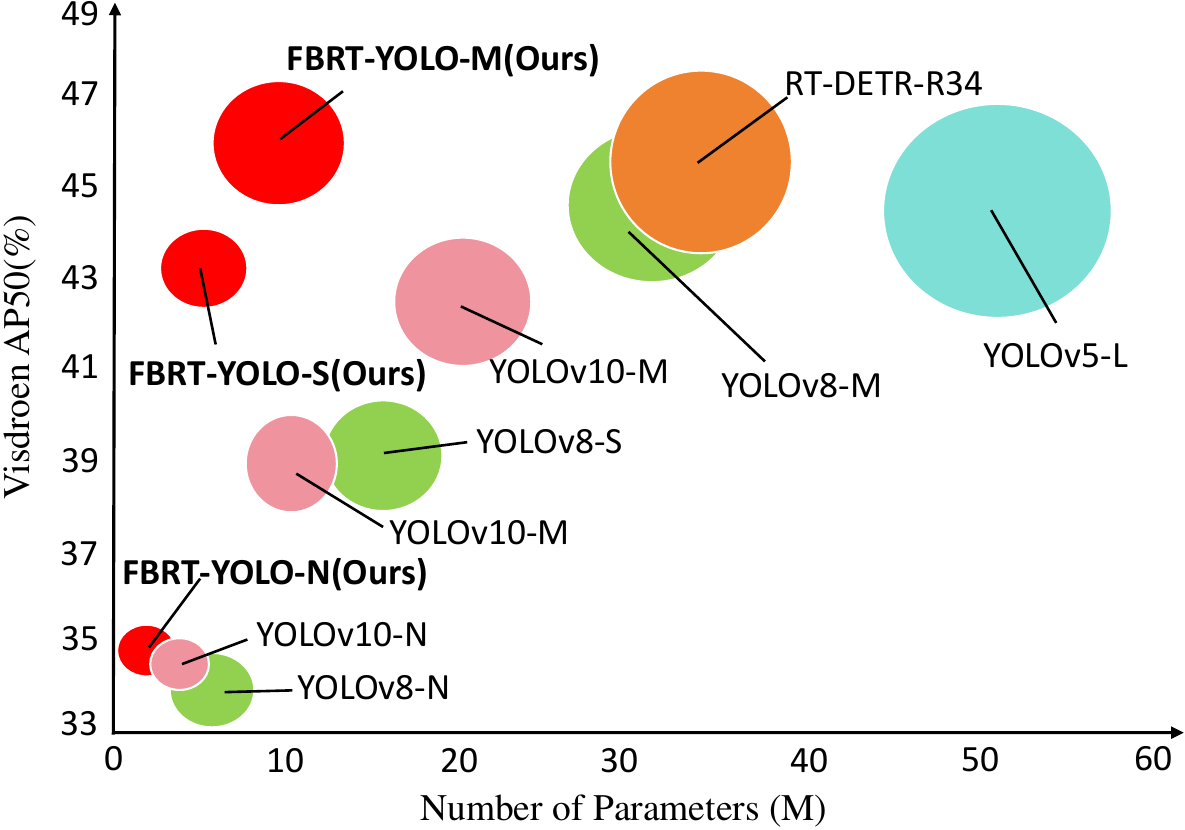}
\caption{Our FBRT-YOLO is compared with other real-time detectors in terms of accuracy and efficiency on VisDrone dataset. The radius of the circle represents GFLOPs.}
\label{fg1}
\end{figure}

Secondly, due to the minimal representation of small objects in aerial images, which often comprise just a few pixels, these objects are susceptible to feature disappearance during convolutional neural network (CNN) feature extraction. To fully utilize the limited feature information and enhance the network’s perception of targets at different scales, we investigate the network's receptive field and propose a Multi-Kernel Perception Unit (MKP). MKP consists of convolutional kernels of different sizes and incorporates spatial point convolutions between these sizes to focus on details at various scales and highlight multi-scale feature representation. We replace the final downsampling layer of the network with MKP. This approach enables multi-scale perception of targets, improving the network's ability to capture features across different scales while further simplifying the network structure.

To meet the requirements of real-time detection in aerial images, we propose FBRT-YOLO, which boasts fewer training parameters and reduced computational load compared to the baseline YOLOv8 model~\shortcite{Jocher_Ultralytics_YOLO_2023}. Extensive experiments conducted on widely-used aerial image benchmarks such as VisDrone~\shortcite{visdrone}, UAVDT~\shortcite{UAV}, and AI-TOD~\shortcite{wang2021tiny} demonstrate that our FBRT-YOLO significantly outperforms previous state-of-the-art YOLO series models in terms of the trade-off between computation and accuracy across various model scales. The results are displayed in \cref{fg1}.
Our contributions can be summarized as follows:
\begin{itemize}
\item We introduce a new family of real-time detectors for aerial image detection across different model scales, named FBRT-YOLO, achieving a highly balanced trade-off between accuracy and efficiency.
\item We propose a Feature Complementary Mapping Module (FCM) that enhances feature matching for small targets in deep networks by integrating rich semantic information with precise spatial positional information.
\item We introduce Multi-Kernel Perception Unit (MKP) to replace the final downsampling operation, enhancing multi-scale target perception,  and simplifying the network for high efficiency.
\end{itemize}

\begin{figure*}[t]
\centering
\includegraphics[width=1\textwidth]{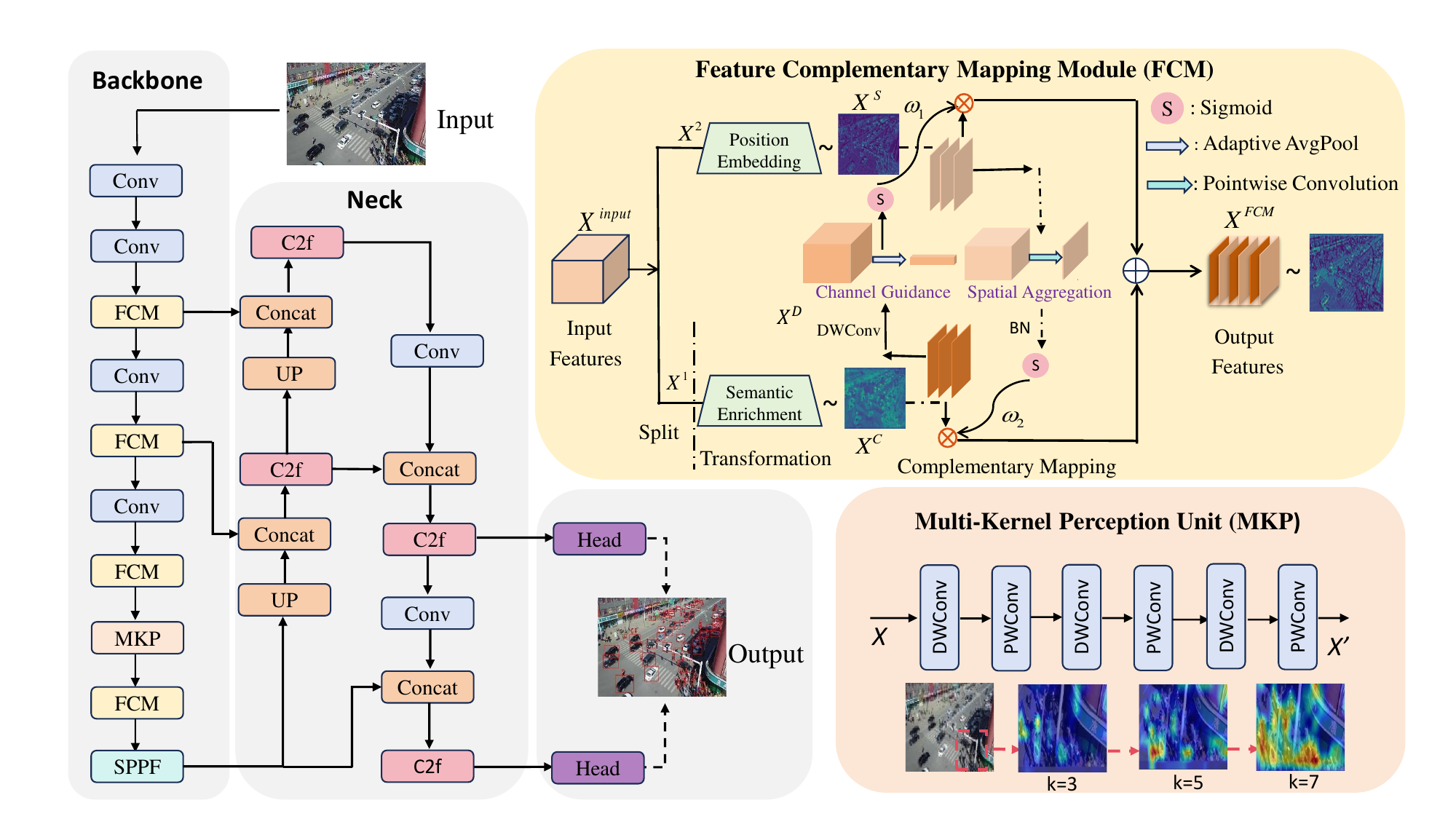} 
\caption{Framework of FBRT-YOLO. FCM module is embedded into each stage of the backbone network to integrate spatial positional information into deeper semantic information. In the final (fourth) stage of the backbone network, MKP units are introduced along with multi-scale convolutions to  enhance perception of targets at various scales. It's worth noting that MKP replaces the final downsampling layer while also reducing the corresponding detection heads.}
\label{fg2}
\end{figure*}

\section{Related Work}
\subsubsection{Real-time Object Detectors.}
Real-time object detectors are crucial for resource-constrained platforms, emphasizing model size, memory, and computational efficiency.
Currently, YOLO~\shortcite{redmon2016you} and FCOS~\shortcite{tian2020fcos} are mainstream frameworks for state-of-the-art real-time object detection. While existing real-time detectors have shown significant performance improvements on public benchmarks such as COCO~\shortcite{COCO,zhao2024detrs}           for low-resolution natural images, their performance on high-resolution aerial images remains unsatisfactory. 
We introduce FBRT-YOLO, a specialized real-time object detector designed to excel in high-resolution aerial settings, demonstrating superior performance compared to existing models.

\subsubsection{Small Object Detection.}
Detecting small objects has long been challenging. Recent solutions include augmenting small object datasets~\shortcite{kisantal2019augmentation} and using high-resolution images to retain detailed features. However, these methods often result in more complex models and slower detection speeds.
ClusDet~\shortcite{yang2019clustered} employs a cluster-based object scale estimation network to effectively detect small objects. DMNet~\shortcite{li2020density} utilizes a density map-based cropping method to leverage spatial and contextual information among objects for improved detection performance. Despite their effectiveness in small object detection, these methods suffer from long inference times and low detection efficiency.
QueryDet~\shortcite{querydet}, while leveraging high-resolution features, incorporates a novel query mechanism to accelerate inference speed for object detectors based on feature pyramids. CEASC~\shortcite{CEASC} introduces a context-enhanced sparse convolution to capture global information and enhance focal features, striking a balance between detection accuracy and efficiency.
These works propose lightweight decoupled heads that to some extent accelerate networks. However, achieving real-time detection remains challenging.
\subsubsection{Multi-scale Information Extraction and Representation.} Small objects are often represented in feature maps by only a few pixels, necessitating multi-scale information to enhance the feature representation of these small objects.
Many works have also been carried out from this aspect~\shortcite{jianan3,jianan2}.
Feature Pyramid Network (FPN) integrates the deep features with the richest semantics information and the shallow features with spatial location information, which alleviates the problem of feature imbalance to a certain extent. 
PANet~\shortcite{panet} adds a bottom-up path on the basis of FPN, which promotes the propagation of bottom-layer information and enhances information exchange. IPG-Net~\shortcite{liu2020ipg} introduces image pyramid into the backbone network to solve the problem of information imbalance. 
The whole process consumes a lot of computing resources, which is not conducive to real-time detection. 
In our work, we focus on integrating deep semantic information with shallow spatial positional information in the backbone network. This integration alleviates the imbalance in information extraction during feature extraction, thereby enhancing the representation of small objects. 
We employ multi-scale convolutional kernels to strengthen the feature representation of targets across various scales.



\section{Method}
We present the entire structure of FBRT-YOLO in \cref{fg2}.
This includes two core lightweight modules: the Feature Complementary Mapping Module and the Multi-Kernel Perception Unit.
FCM aims to integrate more spatial positional information into rich semantic features, enhancing the representation of small objects. MKP utilizes diverse convolutional kernels to capture target information across multiple scales. Additionally, for aerial image detection, we streamline the baseline network by removing non-critical or redundant computations, further refining the network.
\subsection{Feature Complementary Mapping Module}
Insufficient integration of spatial positional and semantic information can lead to mismatches and misalignments in target information. To address this limitation, we propose the Feature Complementary Mapping Module. This module implicitly encodes more low-level spatial information into high-dimensional vectors, transmitting it to deeper layers of the network. This enables the detector to capture stronger structural information, thereby enhancing the expression of semantic information.
The detailed structure of FCM is shown in \cref{fg2}, which utilizes a split, transformation, complementary mapping strategy
and feature aggregation.
The following is a detailed introduction to this module.
\subsubsection{Channel Split.}
We first split the channels of the input feature($X^{\textup{input}}\in \mathbb{R}^{C\times H\times W}$) into two parts with $\alpha C $ channels and $(1-\alpha) C $ channels, where 0$\le  \alpha  \le$ 1 is the split ratio.
The value of $\alpha$ is quite important in the network.
As the network deepens, the branch with lower-level spatial information becomes more prominent, with increasing amounts of low-level spatial information being implicitly encoded into high-dimensional vectors. Enhancing the acquisition of low-level information at appropriate times can improve performance.
The split stage can be formulated as:
\begin{equation}
    (X^1,X^2)=\textup{Split}(X^{\textup{input}}),
  \label{eq:important}
\end{equation}
where $X^{1}\in \mathbb{R}^{\alpha C\times H\times W},X^{2}\in \mathbb{R}^{(1-\alpha) C\times H\times W}$.

\subsubsection{Orientation Transformation.}
To separately obtain spatial mappings of semantic and positional information, we send the obtained $X^1$ to the branch composed of standard $3\times3$ convolution, more rich feature information is extracted on each channel, it is represented as $X^C$ in \cref{fg2}.
$X^2$  is sent to the branch composed of point-wise convolution, the point-wise convolution extracts relatively weak information, preserving a large amount of shallow spatial position information, it is represented as $X^S$.
This transformation process is represented by the formula:
\begin{equation}
   (X^C,X^S) = \phi_1(X^1,X^2),
  \label{eq:important}
\end{equation}
where $\phi_1$ represents learning a mapping relationship between spatial and semantic information, $ X^{C}\in \mathbb{R}^{ C\times H\times W} $ contains rich channel information, $X^{S}\in \mathbb{R}^{ C\times H\times W}$ retains more original spatial location information.

\begin{table*}[h!]
\centering
\footnotesize
\renewcommand{\arraystretch}{1}
\setlength{\tabcolsep}{24pt}
\begin{tabular}{l|cc|c|c|c}
\toprule
\textbf{Model} & \textbf{AP} & $\textbf{AP}_{50}$& \textbf{FLOPs} & \textbf{Params}  & \textbf{FPS} \\
\midrule
\raggedright
YOLOv5-L ~\shortcite{Jocher_YOLOv5_by_Ultralytics_2020} & 27.3 & 44.5 & 109.1 G & 46.5 M& 65 \\
\midrule
\raggedright
YOLOv8-N ~\shortcite{Jocher_Ultralytics_YOLO_2023} & 19.6 & 33.3 & 8.7 G & 3.2 M& 174 \\
\raggedright
YOLOv8-S ~\shortcite{Jocher_Ultralytics_YOLO_2023} & 23.6 & 39.6 & 28.6 G & 11.2 M& 134 \\
\raggedright
YOLOv8-M ~\shortcite{Jocher_Ultralytics_YOLO_2023} & 27.1 & 44.4 & 78.9 G & 25.9 M& 83 \\
\raggedright
YOLOv8-L ~\shortcite{Jocher_Ultralytics_YOLO_2023} & 28.4 & 45.9 & 165.2 G& 43.7 M& 61 \\
\raggedright
YOLOv8-X ~\shortcite{Jocher_Ultralytics_YOLO_2023} & 28.9 & 46.8 & 257.8 G& 68.2 M& 47 \\
\midrule
\raggedright
YOLOv9-M ~\shortcite{wang2024yolov9} & 27.2 & 44.6 & 76.3 G& 20.0 M& 77 \\
\midrule
\raggedright
YOLOv10-S ~\shortcite{wang2024yolov10} & 23.8 & 39.3 & 21.6 G& 7.2 M& 135 \\
\raggedright
YOLOv10-L ~\shortcite{wang2024yolov10} & 27.6 & 44.6 & 120.3 G& 24.4 M& 64 \\
\raggedright
YOLOv10-X ~\shortcite{wang2024yolov10} & 28.7 & 46.1 & 160.4 G& 29.5 M& 49 \\
\midrule
\raggedright
RT-DETR-R34~\shortcite{zhao2024detrs} & 27.2 & 46.0 & 92.0 G& 31.0 M& 88 \\
\raggedright
RT-DETR-R50 ~\shortcite{zhao2024detrs} & 28.8 & 48.3 & 136.0 G& 42.0 M& 65 \\
\midrule
\raggedright
FBRT-YOLO-N (Ours) & 20.2 & 34.4 & 6.7 G& 0.9 M& 192 \\
\raggedright
FBRT-YOLO-S (Ours) & 25.9 & 42.4 & 22.9 G& 2.9 M& 143 \\
\raggedright
FBRT-YOLO-M (Ours) & 28.4 & 45.9 & 58.7 G& 7.2 M& 94 \\
\raggedright
FBRT-YOLO-L (Ours) & 29.7 & 47.7 & 119.2 G& 14.6 M& 70 \\
\raggedright
FBRT-YOLO-X (Ours) & 30.1 & 48.4 & 185.8 G& 22.8 M& 52 \\
\bottomrule
\end{tabular}
\caption{Comparison of AP(\%) and Params/FPS on VisDrone by using our methods with different real-time object detectors.}
\label{t1}
\end{table*}

\begin{table}
\begin{center}
\footnotesize
\renewcommand{\arraystretch}{1}
\setlength{\tabcolsep}{12pt} 
\begin{tabular}{l|ccc}
\toprule
\textbf{Model}   & \textbf{AP} & $\textbf{AP}_{50}$ & $\textbf{AP}_{75}$  \\
\midrule
\raggedright
YOLOv3-SPP3~\shortcite{zhang2019slimyolov3} & 26.4 & - & - \\
\raggedright
YOLOv3-SPP3~\shortcite{zhang2019slimyolov3} & 26.4 & - & - \\
\raggedright
\raggedright
DMNet ~\shortcite{li2020density}  & 29.4 & 49.3 & 30.6 \\
\raggedright
QueryDet ~\shortcite{querydet} & 28.3 & 48.1 & 28.8 \\
\raggedright
CEASC  ~\shortcite{CEASC}  & 28.7 & 50.7 & 28.4 \\
\raggedright
YOLC ~\shortcite{liu2024yolc}  & 28.9 & 51.4 & 28.3 \\
\midrule
\raggedright
FBRT-YOLO  (Ours) & \textbf{30.1} & 48.4 & \textbf{31.7} \\
\bottomrule
\end{tabular}
\end{center}
\caption{Comparison of AP(\%) with state-of-the-art detectors on VisDrone.}
\label{t2}
\end{table}



\subsubsection{Complementary Mapping.}
Currently, the features we obtain, $X^C$ and $X^S$, while effective, are discrete. This can lead to imprecise matching of target features. Therefore, we perform complementary mapping between them to compensate for their respective missing feature mappings, achieving efficient feature matching.
We take $X^C$, which has richer channel information, into channel interaction. It can assign unique weights to the important information on each channel. This is then mapped to $X^S$, which has low-level spatial location information features, for complementary feature fusion. This allows the information after interaction to obtain higher-level features.
Similarly,  $X^S$ with richer low-level spatial position information, through spatial interaction, assigns unique weights to the important information on each position, and maps it to $X^C$ with rich channel information features, to achieve complementary integration and obtain higher-level features.
This process achieves the guidance of stronger features to guide weaker ones, thereby alleviating the problem of information imbalance.

Channel Interaction: First, we use a the depthwise convolution to perform convolution operations on each channel, cutting off the information between the channels, which is calculated as:
\begin{equation}
   X_i^{D} = \phi_2(k_i,X_i^C),
  \label{eq:important}
\end{equation}
where $\phi_2$ represents the mapping of each feature layer channel, $k_i$ is the i-th convolution kernel, $X_i^C$ is the i-th input channel, $X_i^{D}$ is the corresponding single output channel. The output result after depthwise convolution is $ X^{D}\in \mathbb{R}^{ C\times H\times W} $.

then perform global average pooling to obtain the global information on each channel, and finally obtain the key information weights through the sigmoid layer.
The unique weight $ \omega _1\in \mathbb{R}^{ C\times 1 \times 1} $ generated on the channel can be represented as:
\begin{equation}
    \omega _1=\mathcal{R}\left (   \frac{1}{H\times W}\sum_{i=0}^{H}\sum_{j=0}^{W}X^{D}(i,j)  \right ),
  \label{eq:important}
\end{equation}
where $\mathcal{R}$ represents an activation function.

Spatial interaction: To further aggregate spatial information, we adopt a simple design, as shown in \cref{fg2}, which consists of  a $1\times1$ spatial convolution layer, BN~\shortcite{ioffe2015batch} and sigmoid.
Finally, we generate a spatial attention map, which is similar to channel interaction, and map it to the branch that go through the $3\times3$ standard convolution, making it more focused on spatial information.The spatial information weight $ \omega _2$ generated can be calculated as:
\begin{equation}
  \omega _2=\mathcal{R}(\mathcal{F}  ( X^S )),
  \label{eq:important}
\end{equation}
where  $\mathcal{F}$ represents convolution mapping with spatial aggregation, $ \omega _2\in \mathbb{R}^{ 1\times H \times W} $.

\subsubsection{Feature Aggregation.}
After obtaining the channel information weight $\omega _1$ and spatial information weight $\omega _2$, they are respectively mapped to the features containing $X^S$ and $X^C$. Then the two branches are connected together to obtain the feature $X^{FCM}$, which  contains features with dual mappings of spatial and semantic relationships. $X^{FCM}$ is calculated as:
\begin{equation}
  X^{FCM} =(X^C\otimes  \omega _2) \oplus (X^S\otimes \omega _1),
  \label{eq:important}
\end{equation}
where $\otimes$ is element-wise multiplication.

Overall, the FCM module adopts an information complementary fusion method with relatively low computational resources. It propagates shallow-level spatial positional information into deeper layers of the network, alleviating the loss of object spatial location information in the backbone network downsampling process.

\subsection{Multi-Kernel Perception Unit}
Small targets in aerial images are often obscured by background noise, resulting in limited effective information. To fully leverage the available feature information, we employ a multi-kernel perception unit to detect targets at different scales and establish spatial relationships across these scales, thereby enhancing the feature representation of contextual and small target information.
As illustrated in \cref{fg2}, the Multi-Kernel Perception Unit (MKP) concatenates convolutional kernels of various sizes sequentially, incorporating point-wise convolutions between kernels of different scales. The entire process can be mathematically represented as follows:
\begin{equation}
X' = \mathcal{T}_{2k+1}(\mathcal{A}(\cdots \mathcal{A}(\mathcal{T}_k(X))\cdots)),
  \label{eq:important}
\end{equation}
where $X$ represents local features of the input, while $X^{'}$ represents globally mapped features across multiple scales.
$\mathcal{T}_{k}$ represents depthwise convolution with kernel size $k$. In our experiment, we set $k=3$. $\mathcal{A}$ represents point-wise convolution transformation.

\subsection{Targeted Reduction of Redundancy-Driven Network Design}
Currently, real-time detection models are primarily designed for traditional low-resolution image detection, but this does not apply well to high-resolution aerial image detection, resulting in significant structural redundancy. For spatial downsampling in feature extraction, channel expansion precedes depthwise convolution sampling~\shortcite{wang2024yolov10}. Post depthwise convolution, there is interference between channels, leading to a loss of spatial information, which is disadvantageous for detecting aerial images in complex environments. However, we decouple this process by first applying group convolutions for spatial downsampling and then using point convolutions for channel expansion. The parameter calculations for both approaches are as follows:
\begin{equation}
  P^{'} = 3\times3 \times C_1 \times C_2,
  \label{eq:important}
\end{equation}

\begin{equation}
  P = 3\times3 \times C_1 \times \frac{ C_1 }{g} + 1\times1 \times C_1 \times C_2,
  \label{eq:important}
\end{equation}
where $P^{'}$ represents the parameter count for standard convolution, and $P$ represents the parameter count for our method. $C_1$ and $C_2$ denote the input and output channel numbers, respectively. During network downsampling, the channel expansion typically results in $C_2 = 2C_1$. $g$ represents the number of groups.

\begin{figure*}[t]
\centering
\includegraphics[width=1\textwidth]{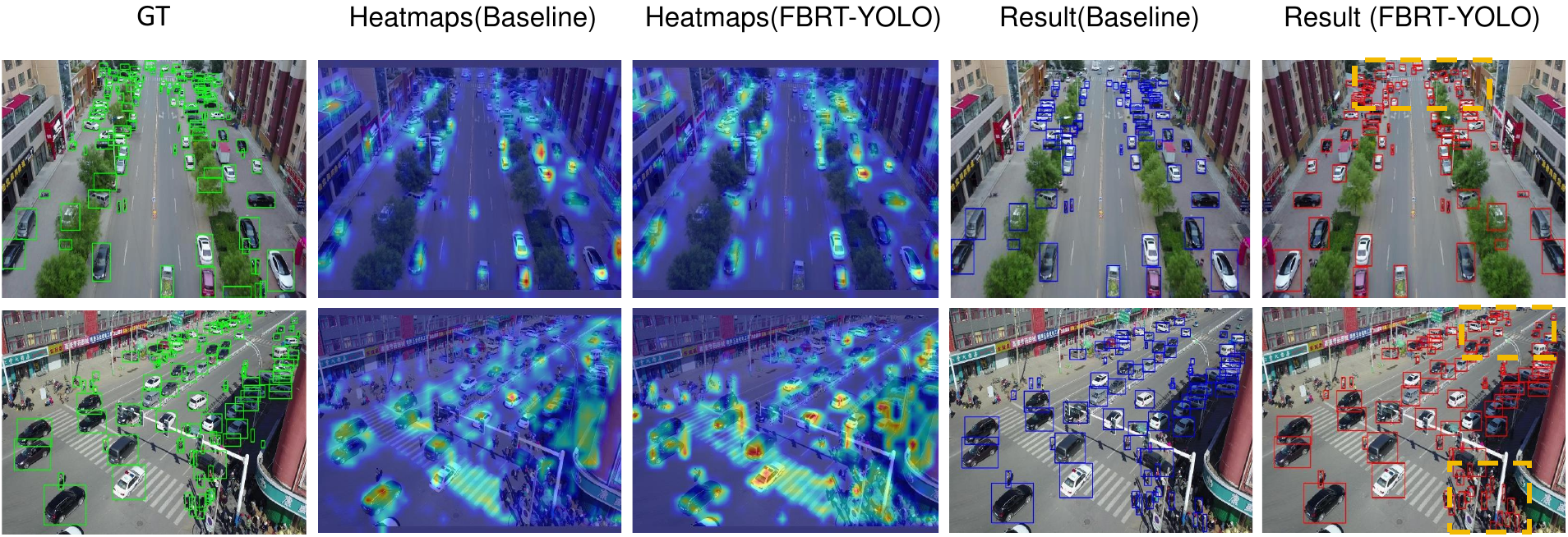} 
\caption{Visualization of the detection results and heatmaps on VisDrone. The highlighted areas represent the regions that the network is focusing on.}
\label{fig4}
\end{figure*}


\section{Experiments}






\subsection{Implementation Details}  
We conduct extensive experiments on three object detection benchmarks based on aerial images, $i.e$. Visdrone, UAVDT, and AI-TOD.
All experiments are conducted on an NVIDIA GeForce RTX 4090 GPU, except that the inference speed is test on a single RTX 3080 GPU.
Our network is trained for 300 epochs using the stochastic gradient descent (SGD) optimizer with a momentum of 0.937, a weight decay of 0.0005, a batch size of 4, and an initial learning rate of 0.01.

\subsection{Results on Visdrone Dataset}
\subsubsection{State-of-the-art Comparison.}
As shown in \cref{t1}, we compare FBRT-YOLO with existing real-time detectors. Our FBRT-YOLO achieves superior performance and faster detection efficiency across various model scales.
For resource-limited aerial operation equipment, we demonstrate results of FBRT-YOLO models at various scales compared to other real-time state-of-the-art object detectors.
For small models, FBRT-YOLO-N/S reduces parameter count by $72\%$ and $74\%$ respectively compared to YOLOv8-N/S, while achieving an improved detection accuracy of $0.6\%$ and $2.3\%$ in average precision (AP).
For medium models, FBRT-YOLO-M reduce GFLOPs by $26\%$ and $23\%$ compared to YOLOv8-M and YOLOv9-M, respectively, while achieving improvements in AP of $1.3\%$ and $1.2\%$, respectively.
For large models, compared to YOLOv8-X and YOLOv10-X, our FBRT-YOLO-X shows $66\%$ and $23\%$ fewer parameters, respectively, and achieves a significant improvement in AP of $1.2\%$ and $1.4\%$.
Moreover, compared to RT-DETR-R34/R50, FBRT-YOLO-M/L achieves fewer parameters, lower GFLOPs, higher detection speed, and better detection performance. These experimental results demonstrate the superiority of our FBRT-YOLO as a real-time aerial image detector.

As shown in \cref{t2}, it shows the comparison results of our method with other state-of-the-art methods on VisDrone. which indicates that our FBRT-YOLO can effectively detects aerial images.

\begin{table}[t]
\begin{center}
\footnotesize
\renewcommand{\arraystretch}{0.95}
\setlength{\tabcolsep}{13pt}
\begin{tabular}{l|ccc}
\toprule
\textbf{Model} & \textbf{AP} & $\textbf{AP}_{50}$ & $\textbf{AP}_{75}$ \\
\midrule
\raggedright
ClusDet ~\shortcite{yang2019clustered} & 13.7 & 26.5 & 12.5 \\
\raggedright
GLSAN ~\shortcite{deng2020global} & 17.0 & 28.1 & 18.8 \\
\raggedright
DREN ~\shortcite{9022557} & 15.1 & - & - \\
\raggedright
GFL ~\shortcite{li2020generalized} & 16.9 & 29.5 & 17.9 \\
\raggedright
CEASC ~\shortcite{CEASC} & 17.1 & 30.9 & 17.8 \\
\midrule
\raggedright
FBRT-YOLO (Ours) & \textbf{18.4} & \textbf{31.1} & \textbf{18.9} \\
\bottomrule
\end{tabular}
\end{center}
\caption{Comparison of AP(\%) with state-of-the-art detectors on UAVDT.}
\label{t3}
\end{table}


\begin{table}[t]
\begin{center}
\footnotesize
\renewcommand{\arraystretch}{1}
\setlength{\tabcolsep}{6pt} 
\begin{tabular}{c|cc|ccc}
\toprule
\textbf{Model}    & \textbf{AP} & $\textbf{AP}_{50}$  & \textbf{Params}&\textbf{FLOPs} & \textbf{FPS}  \\
\midrule
YOLOv8-S  & 19.1  & 43.6  & 11.2 M &28.6 G& 131  \\
FBRT-YOLO-S   & \textbf{20.2} &  \textbf{45.8}& \textbf{2.9 M}& \textbf{22.9 G}& \textbf{142}  \\
\bottomrule
\end{tabular}
\end{center}
\caption{ Comparison of AP(\%) and Params/FPS on AI-TOD by using our methods with baseline.}
\label{t4}
\end{table}

\begin{table}[t]
\begin{center}
\footnotesize
\renewcommand{\arraystretch}{1}
\setlength{\tabcolsep}{6.5pt} 
\begin{tabular}{ccc|cc|cc}
\toprule
\textbf{RR}& \textbf{FCM}  & \textbf{MKP} & \textbf{AP}&$\textbf{AP}_{50}$& \textbf{Params} & \textbf{FLOPs} \\
\midrule
 \ding{55} &\ding{55}  &  \ding{55} & 23.6 &39.6 & 11.2 M &28.6 G  \\
\checkmark  &  \ding{55} &  \ding{55}& 23.4 &39.2 & 9.1 M & 25.5 G\\
 \checkmark & \checkmark &  \ding{55} & 24.3 &40.6& 7.2 M  & 23.2 G \\
\checkmark  &\checkmark  &\checkmark  & \textbf{25.9}  &\textbf{42.4} &\textbf{2.9 M} &\textbf{22.9 G}\\
\bottomrule
\end{tabular}
\end{center}
\caption{ Ablation on FCM, MKP and RR Module on VisDrone. 'RR' represents operations aimed at reducing inherent redundancy in the network. Replace the final downsampling layer with MKP and remove the corresponding detection head.
}
\label{t5}
\end{table}

\subsubsection{Qaulitative Resutls.}
To better demonstrate the superior performance of FBRT-YOLO in detecting aerial images, we visualize the heatmaps of both the baseline model and our method in \cref{fig4}. From the results, we observe that FBRT-YOLO enhances focus on small and densely packed targets, showcasing the method's superiority in enhancing spatial and multiscale information within the network.

\subsection{Results on UAVDT Dataset}
\subsubsection{Quantitative Result.}
\cref{t3} reports our comparison results on the UAVDT dataset. Our proposed method surpasses existing methods, such as GLSAN~\shortcite{deng2020global} and CEASC~\shortcite{CEASC}. The results clearly show that our proposed FBRT-YOLO achieves superior performance with an AP of $18.4\%$, outperforming other state-of-the-art methods in aerial image detection. This demonstrates the effectiveness of our detection framework.

\subsubsection{Qaulitative Resutls.}
\begin{figure}[t]
\centering
\includegraphics[width=0.47\textwidth]{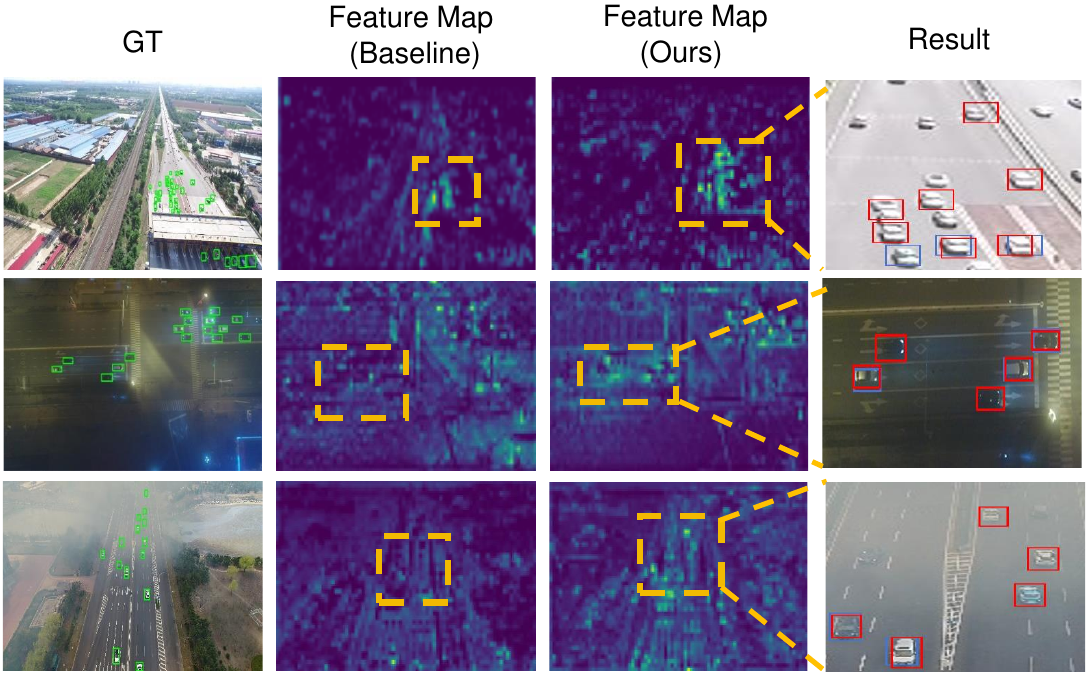} 
\caption{Visualizations of the detection results of baseline and our proposed method under low light and similar background conditions on UAVDT. The blue boxes represent the prediction results using the baseline model, while the red boxes represent the prediction results using our method.}
\label{fig5}
\end{figure}

A complex background can significantly limit the effective information about the target. Our method focuses on effectively propagating the spatial information of the target through network layers to enhance feature representation. Visualization of detection results, as shown in \cref{fig5}, proves that our method significantly improves detection performance in complex backgrounds.

\subsection{Results on AI-TOD Dataset}

The AI-TOD dataset contains a significant proportion of small objects. To better validate the superiority of our method in small object detection, we also evaluate FBRT-YOLO on AI-TOD. As reported in \cref{t4}, our method reduces the parameter count by $74\%$, the GFLOPs by $20\%$, while achieving a $2.2\%$ increase in $\text{AP}_{50}$ and a $1.1\%$ increase in AP compared to the baseline.
\subsection{Ablation Study}
To validate the effectiveness of the core module design in FBRT-YOLO, we design a series of ablation experiments on the VisDrone dataset. We use YOLOv8-S as the baseline model in all the ablation experiments.
\subsubsection{Effect of Key Components.} 
Experimental results in \cref{t5} exhibit the effectiveness of all contributions in this work. 
We reduce inherent redundancy in the baseline model, optimize it, and achieve a $18\%$ reduction in parameters and a $11\%$ decrease in computational load, albeit with a slight decrease in accuracy. Introducing the FCM module into various stages of the backbone network incorporates spatial positional information in deeper layers, resulting in a $1.4\%$ increase in $\text{AP}_{50}$ and further reducing network computational resources. We replace the downsampling operation of the backbone network's final layer with MKP units to detect targets at multiple scales, thereby increasing AP by $1.6\%$.
It’s worth noting that our network converges faster during the training process compared to the baseline network.

\begin{table}[t]
\begin{center}
\footnotesize
\renewcommand{\arraystretch}{1}
\setlength{\tabcolsep}{8pt} 
\begin{tabular}{cc|cc}
\toprule
\textbf{Channel Mapping}    & \textbf{Spatial Mapping} & \textbf{AP}  & $\textbf{AP}_{50}$   \\
\midrule
  \ding{55} &  \ding{55} & 25.0  &40.4     \\
\checkmark  &  \ding{55} & 25.6  &41.7  \\
 \ding{55}  & \checkmark  & 25.3  &41.3     \\
\checkmark  & \checkmark  & \textbf{25.9}  &\textbf{42.4}    \\
\bottomrule
\end{tabular}
\end{center}
\caption{ The experiment validates the optimal configuration of the mapping relationship.}
\label{t6}
\end{table}

\subsubsection{Effect of Mapping Relationship.} 
\cref{t6} shows the results of the proposed channel and spatial complementary mapping.
In order to obtain the optimal configuration of two mapping relationships, we design a series of variant experiments.
According to the experimental results, we find that models using channel or spatial mapping are superior to models without mapping relationships.
Combining the two can achieve better results.  
Compared with the model without mapping relationships, this optimal configuration has improved $\text{AP}_{50}$  by $2.0\%$.
\begin{table}[t]
\begin{center}
\footnotesize
\renewcommand{\arraystretch}{1}
\setlength{\tabcolsep}{11pt} 
\begin{tabular}{c|cc|c}
\toprule
 \textbf{Split Ratio ($\alpha$)}    & \textbf{AP} & $\textbf{AP}_{50}$  & \textbf{Params}    \\
\midrule
(0.5,0.5,0.5,0.5)  & 25.7  &42.2   & 3.2 M   \\
(0.25,0.25,0.25,0.25) & 25.5  &42.1   &\textbf{2.6 M}    \\
(0.75,0.75,0.75,0.75) & 25.8  &42.3   & 4.0 M    \\
(0.75,0.75,0.25,0.25)& \textbf{25.9}  &\textbf{42.4}   &2.9 M     \\
\bottomrule
\end{tabular}
\end{center}
\caption{Experimental verification of the impact of the spatial and channel feature partition ratio of $\alpha$ in the FCM module at each stage of the backbone network.}
\label{t7}
\end{table}
\subsubsection{Effect of Split Ratio.} 
\cref{t7} shows the impact of different parameters $\alpha$ on the experimental results, where $\alpha$ represents the split ratio of spatial feature information and channel feature information.
From the experimental results, we can see that as the downsampling process progresses, the proportion of the spatial feature part (undergoing point-wise convolution) increases, and the experimental effect will be better.
We speculate that the reason for this phenomenon is that when $\alpha$ takes the values 0.75, 0.75, 0.25, 0.25, it retains more spatial location information in deeper networks, which is beneficial for the localization and matching of target features.
Retaining more spatial position information in a deeper network is also consistent with the original intention of the FCM module design.

\begin{table}[t]
\begin{center}
\footnotesize
\renewcommand{\arraystretch}{1}
\setlength{\tabcolsep}{9pt} 
\begin{tabular}{c|cc|cc}
\toprule
\textbf{Kernel Size}    & \textbf{AP} & $\textbf{AP}_{50}$  & \textbf{Params} & \textbf{FLOPs}  \\
\midrule
(3,3,3)  & 25.4  &41.7  & \textbf{2.82 M} & \textbf{22.7 G}  \\
(5,5,5)  & 25.6  &42.0  & 2.88 M & 22.8 G  \\
(7,7,7)  & 25.8  &42.2 & 2.94 M & 23.1 G  \\
(3,5,7)  & \textbf{25.9}  &\textbf{42.4}  &2.90 M  &  22.9 G \\
\bottomrule
\end{tabular}
\end{center}
\caption{ Experiments with different kernel size in MKP.}
\label{t8}
\end{table}
\subsubsection{Effect of Kernel Size.} 
\cref{t8} shows the experimental results of different kernel sizes in MKP. From the experimental results, it can be observed that smaller kernels provide limited receptive fields for the network, failing to establish strong contextual associations, while larger kernels introduce significant background noise, which is detrimental to detection. By using convolutional kernels of varying sizes, we capture multi-scale features of targets spanning different sizes. Additionally, we introduce point-wise convolutions between different kernel sizes to integrate spatial information across scales, achieving optimal performance.

\section{Conclusion}
In this paper, we propose a new family of real-time detectors for aerial image detection, named FBRT-YOLO. Specifically, it introduces two lightweight modules: the Feature Complementary Mapping Module (FCM), which aims to improve the fusion of rich semantic information with precise spatial location details, and the Multi-Kernel Perception Unit (MKP), which enhances multi-scale target perception and improves the network's ability to capture features across varying scales. For aerial image detection, we also reduce the inherent redundancies in conventional detectors, further accelerating the network. Extensive experimental results on the VisDrone, UAVDT, and AI-DOT datasets demonstrate that FBRT-YOLO achieves a highly balanced trade-off between accuracy and efficiency in aerial image detection.

\section{Acknowledgements}
This work was supported by Natural Science Foundation of Chongqing, China (Grant No. cstc2021jcyj-msxmX1130).

\bibliography{aaai25}

\begin{thebibliography}{31}
\providecommand{\natexlab}[1]{#1}

\bibitem[{Deng et~al.(2020)Deng, Li, Xie, Song, Liao, Hao, and Qin}]{deng2020global}
Deng, S.; Li, S.; Xie, K.; Song, W.; Liao, X.; Hao, A.; and Qin, H. 2020.
\newblock A global-local self-adaptive network for drone-view object detection.
\newblock \emph{IEEE Transactions on Image Processing}, 30: 1556--1569.

\bibitem[{Du et~al.(2023)Du, Huang, Chen, and Huang}]{CEASC}
Du, B.; Huang, Y.; Chen, J.; and Huang, D. 2023.
\newblock Adaptive sparse convolutional networks with global context enhancement for faster object detection on drone images.
\newblock In \emph{Proceedings of the IEEE/CVF Conference on Computer Vision and Pattern Recognition}, 13435--13444.

\bibitem[{Du et~al.(2018)Du, Qi, Yu, Yang, Duan, Li, Zhang, Huang, and Tian}]{UAV}
Du, D.; Qi, Y.; Yu, H.; Yang, Y.; Duan, K.; Li, G.; Zhang, W.; Huang, Q.; and Tian, Q. 2018.
\newblock The unmanned aerial vehicle benchmark: Object detection and tracking.
\newblock In \emph{Proceedings of the European conference on computer vision (ECCV)}, 370--386.

\bibitem[{Huang et~al.(2024)Huang, Li, Chen, Wang, Zhao, and Xu}]{jianan2}
Huang, B.; Li, J.; Chen, J.; Wang, G.; Zhao, J.; and Xu, T. 2024.
\newblock Anti-UAV410: A Thermal Infrared Benchmark and Customized Scheme for Tracking Drones in the Wild.
\newblock \emph{IEEE Transactions on Pattern Analysis and Machine Intelligence}, 46(5): 2852--2865.

\bibitem[{Ioffe and Szegedy(2015)}]{ioffe2015batch}
Ioffe, S.; and Szegedy, C. 2015.
\newblock Batch normalization: Accelerating deep network training by reducing internal covariate shift.
\newblock In \emph{International conference on machine learning}, 448--456. pmlr.

\bibitem[{Jocher(2020)}]{Jocher_YOLOv5_by_Ultralytics_2020}
Jocher, G. 2020.
\newblock {YOLOv5 by Ultralytics}.

\bibitem[{Jocher, Chaurasia, and Qiu(2023)}]{Jocher_Ultralytics_YOLO_2023}
Jocher, G.; Chaurasia, A.; and Qiu, J. 2023.
\newblock {Ultralytics YOLO}.

\bibitem[{Kisantal et~al.(2019)Kisantal, Wojna, Murawski, Naruniec, and Cho}]{kisantal2019augmentation}
Kisantal, M.; Wojna, Z.; Murawski, J.; Naruniec, J.; and Cho, K. 2019.
\newblock Augmentation for small object detection.
\newblock \emph{arXiv preprint arXiv:1902.07296}.

\bibitem[{Li et~al.(2020{\natexlab{a}})Li, Yang, Zhu, Chen, and Guan}]{li2020density}
Li, C.; Yang, T.; Zhu, S.; Chen, C.; and Guan, S. 2020{\natexlab{a}}.
\newblock Density map guided object detection in aerial images.
\newblock In \emph{proceedings of the IEEE/CVF conference on computer vision and pattern recognition workshops}, 190--191.

\bibitem[{Li et~al.(2018)Li, Liang, Shen, Xu, Feng, and Yan}]{jianan3}
Li, J.; Liang, X.; Shen, S.; Xu, T.; Feng, J.; and Yan, S. 2018.
\newblock Scale-Aware Fast R-CNN for Pedestrian Detection.
\newblock \emph{IEEE Transactions on Multimedia}, 20(4): 985--996.

\bibitem[{Li et~al.(2017)Li, Liang, Wei, Xu, Feng, and Yan}]{jianan1}
Li, J.; Liang, X.; Wei, Y.; Xu, T.; Feng, J.; and Yan, S. 2017.
\newblock Perceptual Generative Adversarial Networks for Small Object Detection.
\newblock In \emph{Proceedings of the IEEE Conference on Computer Vision and Pattern Recognition (CVPR)}.

\bibitem[{Li et~al.(2020{\natexlab{b}})Li, Wang, Wu, Chen, Hu, Li, Tang, and Yang}]{li2020generalized}
Li, X.; Wang, W.; Wu, L.; Chen, S.; Hu, X.; Li, J.; Tang, J.; and Yang, J. 2020{\natexlab{b}}.
\newblock Generalized focal loss: Learning qualified and distributed bounding boxes for dense object detection.
\newblock \emph{Advances in Neural Information Processing Systems}, 33: 21002--21012.

\bibitem[{Lin et~al.(2017)Lin, Doll{\'a}r, Girshick, He, Hariharan, and Belongie}]{FPN}
Lin, T.-Y.; Doll{\'a}r, P.; Girshick, R.; He, K.; Hariharan, B.; and Belongie, S. 2017.
\newblock Feature pyramid networks for object detection.
\newblock In \emph{Proceedings of the IEEE conference on computer vision and pattern recognition}, 2117--2125.

\bibitem[{Lin et~al.(2014)Lin, Maire, Belongie, Hays, Perona, Ramanan, Doll{\'a}r, and Zitnick}]{COCO}
Lin, T.-Y.; Maire, M.; Belongie, S.; Hays, J.; Perona, P.; Ramanan, D.; Doll{\'a}r, P.; and Zitnick, C.~L. 2014.
\newblock Microsoft coco: Common objects in context.
\newblock In \emph{Computer Vision--ECCV 2014: 13th European Conference, Zurich, Switzerland, September 6-12, 2014, Proceedings, Part V 13}, 740--755. Springer.

\bibitem[{Liu et~al.(2024)Liu, Gao, Huang, Hu, Liu, and Wang}]{liu2024yolc}
Liu, C.; Gao, G.; Huang, Z.; Hu, Z.; Liu, Q.; and Wang, Y. 2024.
\newblock YOLC: You Only Look Clusters for Tiny Object Detection in Aerial Images.
\newblock \emph{IEEE Transactions on Intelligent Transportation Systems}.

\bibitem[{Liu et~al.(2018)Liu, Qi, Qin, Shi, and Jia}]{panet}
Liu, S.; Qi, L.; Qin, H.; Shi, J.; and Jia, J. 2018.
\newblock Path aggregation network for instance segmentation.
\newblock In \emph{Proceedings of the IEEE conference on computer vision and pattern recognition}, 8759--8768.

\bibitem[{Liu et~al.(2020)Liu, Gao, Sun, and Fang}]{liu2020ipg}
Liu, Z.; Gao, G.; Sun, L.; and Fang, L. 2020.
\newblock IPG-net: Image pyramid guidance network for small object detection.
\newblock In \emph{Proceedings of the IEEE/CVF conference on computer vision and pattern recognition workshops}, 1026--1027.

\bibitem[{Redmon et~al.(2016)Redmon, Divvala, Girshick, and Farhadi}]{redmon2016you}
Redmon, J.; Divvala, S.; Girshick, R.; and Farhadi, A. 2016.
\newblock You only look once: Unified, real-time object detection.
\newblock In \emph{Proceedings of the IEEE conference on computer vision and pattern recognition}, 779--788.

\bibitem[{Tian et~al.(2020)Tian, Shen, Chen, and He}]{tian2020fcos}
Tian, Z.; Shen, C.; Chen, H.; and He, T. 2020.
\newblock FCOS: A simple and strong anchor-free object detector.
\newblock \emph{IEEE transactions on pattern analysis and machine intelligence}, 44(4): 1922--1933.

\bibitem[{Wang et~al.(2024)Wang, Chen, Liu, Chen, Lin, Han, and Ding}]{wang2024yolov10}
Wang, A.; Chen, H.; Liu, L.; Chen, K.; Lin, Z.; Han, J.; and Ding, G. 2024.
\newblock Yolov10: Real-time end-to-end object detection.
\newblock \emph{arXiv preprint arXiv:2405.14458}.

\bibitem[{Wang, Bochkovskiy, and Liao(2023)}]{wang2023yolov7}
Wang, C.-Y.; Bochkovskiy, A.; and Liao, H.-Y.~M. 2023.
\newblock YOLOv7: Trainable bag-of-freebies sets new state-of-the-art for real-time object detectors.
\newblock In \emph{Proceedings of the IEEE/CVF conference on computer vision and pattern recognition}, 7464--7475.

\bibitem[{Wang, Yeh, and Liao(2024)}]{wang2024yolov9}
Wang, C.-Y.; Yeh, I.-H.; and Liao, H.-Y.~M. 2024.
\newblock Yolov9: Learning what you want to learn using programmable gradient information.
\newblock \emph{arXiv preprint arXiv:2402.13616}.

\bibitem[{Wang et~al.(2020)Wang, Sun, Cheng, Jiang, Deng, Zhao, Liu, Mu, Tan, Wang et~al.}]{wang2020deep}
Wang, J.; Sun, K.; Cheng, T.; Jiang, B.; Deng, C.; Zhao, Y.; Liu, D.; Mu, Y.; Tan, M.; Wang, X.; et~al. 2020.
\newblock Deep high-resolution representation learning for visual recognition.
\newblock \emph{IEEE transactions on pattern analysis and machine intelligence}, 43(10): 3349--3364.

\bibitem[{Wang et~al.(2021)Wang, Yang, Guo, Zhang, and Xia}]{wang2021tiny}
Wang, J.; Yang, W.; Guo, H.; Zhang, R.; and Xia, G.-S. 2021.
\newblock Tiny object detection in aerial images.
\newblock In \emph{2020 25th international conference on pattern recognition (ICPR)}, 3791--3798. IEEE.

\bibitem[{Wang et~al.(2022)Wang, Zhang, Yang, and Sun}]{wang2022anchor}
Wang, Y.; Zhang, X.; Yang, T.; and Sun, J. 2022.
\newblock Anchor detr: Query design for transformer-based detector.
\newblock In \emph{Proceedings of the AAAI conference on artificial intelligence}, volume~36, 2567--2575.

\bibitem[{Yang, Huang, and Wang(2022)}]{querydet}
Yang, C.; Huang, Z.; and Wang, N. 2022.
\newblock QueryDet: Cascaded sparse query for accelerating high-resolution small object detection.
\newblock In \emph{Proceedings of the IEEE/CVF Conference on computer vision and pattern recognition}, 13668--13677.

\bibitem[{Yang et~al.(2019)Yang, Fan, Chu, Blasch, and Ling}]{yang2019clustered}
Yang, F.; Fan, H.; Chu, P.; Blasch, E.; and Ling, H. 2019.
\newblock Clustered object detection in aerial images.
\newblock In \emph{Proceedings of the IEEE/CVF international conference on computer vision}, 8311--8320.

\bibitem[{Zhang et~al.(2019)Zhang, Huang, Chen, and Zhang}]{9022557}
Zhang, J.; Huang, J.; Chen, X.; and Zhang, D. 2019.
\newblock How to Fully Exploit The Abilities of Aerial Image Detectors.
\newblock In \emph{2019 IEEE/CVF International Conference on Computer Vision Workshop (ICCVW)}, 1--8.

\bibitem[{Zhang, Zhong, and Li(2019)}]{zhang2019slimyolov3}
Zhang, P.; Zhong, Y.; and Li, X. 2019.
\newblock SlimYOLOv3: Narrower, Faster and Better for Real-Time UAV Applications.
\newblock In \emph{2019 IEEE/CVF International Conference on Computer Vision Workshop (ICCVW)}, 37--45.

\bibitem[{Zhao et~al.(2024)Zhao, Lv, Xu, Wei, Wang, Dang, Liu, and Chen}]{zhao2024detrs}
Zhao, Y.; Lv, W.; Xu, S.; Wei, J.; Wang, G.; Dang, Q.; Liu, Y.; and Chen, J. 2024.
\newblock Detrs beat yolos on real-time object detection.
\newblock In \emph{Proceedings of the IEEE/CVF Conference on Computer Vision and Pattern Recognition}, 16965--16974.

\bibitem[{Zhu et~al.(2018)Zhu, Wen, Bian, Ling, and Hu}]{visdrone}
Zhu, P.; Wen, L.; Bian, X.; Ling, H.; and Hu, Q. 2018.
\newblock Vision meets drones: A challenge.
\newblock \emph{arXiv preprint arXiv:1804.07437}.

\end{thebibliography}

\end{document}